%% file: main.tex
\documentclass{article}

\usepackage{arxiv}
\usepackage{hyperref}

\usepackage{amsmath}
\usepackage{amssymb}
\usepackage{comment}
\usepackage{tikz}
\usetikzlibrary{automata, positioning, arrows, arrows.meta, trees, shapes.geometric, fit, backgrounds}
\usepackage{booktabs}
\usepackage{cleveref}

\title{Learning of Tree Automata applied to Neural Language Acceptors\thanks{Presented as WIP paper at ICGI 2026, Delft, the Netherlands, October 2026}}

\author{
J. {da Silva}\\ Universidad ORT Uruguay\\ \texttt{jd229475@ort.edu.uy}
\And
S. Yovine\href{https://orcid.org/0000-0002-2737-4382}{\includegraphics[scale=0.06]{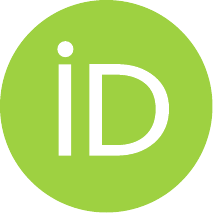}}\\
Universidad ORT Uruguay\\ \texttt{yovine@ort.edu.uy}
}

\begin{document}

\maketitle

\input{src/abstract}


%\tableofcontents

\input{src/intro}
\input{src/background}
\input{src/shortages}

\input{src/leakage}
\input{src/hybrid}
\input{src/experiments}
\input{src/concl}

\paragraph{Acknowledgments}
Partially funded by Agencia Nacional de Investigación e Innovación (ANII), Uruguay, under grants \verb|FMV_1_2023_1_175864| and \verb|POS_NAC_2023_1_178663|, and Programa de Desarrollo de las Ciencias Básicas (PEDECIBA), Uruguay.

%\acks{Acknowledgements go here.}

\bibliographystyle{unsrt}
\bibliography{references}

\end{document}

%% file: src/abstract.tex
\begin{abstract}
We extract visibly pushdown grammars (VPGs) from neural language models in a
fully black-box setting. Our work builds on the VPL\textsuperscript{*} framework of~\cite{barbot}, which learns VPGs from recurrent networks by exploiting access
to the target's internal state.
First, we replace the white-box equivalence oracle
with PAC sampling over trees. The resulting algorithm applies
unchanged to transformer language models and, more generally, to any
acceptor exposing only binary decisions. 
Second, we further observe that the original framework only ever queries the target on well-formed sequences that can be accepted by the most permissive VPG (called BParse) with the target's ranked alphabet.
That is, VPL\textsuperscript{*} is blind to the target's behavior outside BParse. Thus, we propose to sample trees directly instead of sequences and 
%dropping the BParse gate on membership queries, 
%we recover a tree automaton describing the target 
%on \emph{all} trees over the structured alphabet, together with a quantitative
call the target with all the resulting sequences in order to provide an error estimate of how far the target deviates from being a VPG.
Last but not least, we develop a learner that captures a tree automaton describing discovered behaviors of the target outside BParse, which results in a larger learnable space. 
We evaluate the approach on several cases, including transformers trained with Dyck grammars and synthetic targets designed to be outside BParse.
\end{abstract}

%% file: src/intro.tex
\section{Introduction}
\label{sec:intro}

The deployment of neural language models in domains where failures carry
real consequences demands more than predictive accuracy, it requires the
ability to inspect their behavior. A model that cannot be examined cannot be
trusted in safety-critical settings, and post-hoc explanations of
individual predictions do little to characterize a model as a whole.
Grammatical inference explores another route: extract a symbolic
surrogate model that captures the language the target model accepts,
and reason about the surrogate. For models trained on inputs with no 
recursive structure, regular surrogates suffice, and a substantial 
body of work extracts deterministic finite automata from recurrent
networks~\cite{angluin1987,mayr2018,weiss2018}. Many languages of practical
interest; like arithmetic expressions, balanced parentheses or the syntax of 
programming languages; are not regular but exhibit nesting, and the
visibly pushdown languages~\cite{alur2004} are the natural class for them,
strictly above regular.

The state of the art for active learning of visibly pushdown grammars
(VPGs) from neural acceptors is the VPL\textsuperscript{*} framework of
\cite{barbot}, which operates over recurrent networks. Its equivalence
oracle runs an A\textsuperscript{*} search over the network's internal
state, presupposing both white-box access to the model and a recurrent
architecture. Membership queries in their framework go through the 
construction of the most permissive VPG (called BParse) with the target's 
ranked alphabet, called \emph{BParse}, which by definition restricts the 
learner's view to well-matched inputs encoded as BParse-shaped trees. Thus,
the extracted VPG describes the target only on this domain, being silent
about whatever the model does on the rest of the input space.

A complementary line of work learns VPGs directly from sample strings
rather than from neural acceptors. \cite{jia2024vstar} propose V-Star,
which adapts Angluin's L\textsuperscript{*} to actively learn visibly
pushdown grammars, inferring nesting structure from program inputs such
as JSON, XML and S-expressions. Like VPL\textsuperscript{*}, the model V-Star learns is always a
visibly pushdown grammar, hence a VPL describing only well-matched
inputs. Our learner instead extracts a tree automaton over the
structured alphabet, a strictly more expressive object that can capture
target behavior outside any VPL and produces witnesses when the target
accepts ill-matched sequences that no VPG could.

This work makes three contributions over \cite{barbot}. 
First, we replace their white-box equivalence oracle with PAC sampling over trees,
yielding a fully black-box variant of
VPL\textsuperscript{*} that applies to any acceptor exposing only binary
decisions. This allows applying the
resulting learning algorithm to transformer language models, a family for which
the A\textsuperscript{*} oracle originally proposed in VPL\textsuperscript{*} 
is not available. 
Besides, we drop the BParse gate on membership queries and sample trees 
directly from the structured alphabet's language, and measure the errors
incurred by VPL\textsuperscript{*} when avoiding querying the target on 
non-well formed sequences.
Finally, building on the previous step, we develop a learner that 
recovers a tree automaton that describes the target on \emph{all} trees 
over the structured alphabet. When the target is not a VPG, the 
BParse-consistent observation table is projected to a VPG that agrees 
with the one produced by VPL\textsuperscript{*}.
We stress the fact that our implementation of VPL\textsuperscript{*}
is the first publicly available one. 
%the extension to the black-box and raw settings described here is built on top of it. 

%, and report extraction results on Dyck-trained transformers
%  together with a quantitative estimate of how far the target 
% deviates from being a visibly pushdown language;

%% file: src/background.tex
\section{Background}
\label{sec:background}

\subsection{The tree encoding}
\label{sec:tree-encoding}

Fix a structured alphabet $\Sigma = \Sigma_\mathit{push} \cup
\Sigma_\mathit{pop} \cup \Sigma_\mathit{int}$ partitioning $\Sigma$ into
push, pop, and internal letters. The well-formed words $W_\Sigma
\subseteq \Sigma^*$ are those in which every push has a matching pop in
the standard bracket sense. Visibly pushdown languages~\cite{alur2004}
are the subsets of $W_\Sigma$ recognized by visibly pushdown grammars
(VPGs), and we treat them throughout via their tree representation.

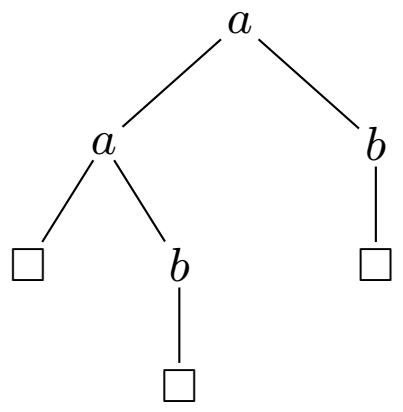
\begin{figure}[t]
\centering
\begin{tikzpicture}[
  level distance=8mm, level 1/.style={sibling distance=18mm},
  level 2/.style={sibling distance=10mm}, level 3/.style={sibling distance=6mm},
  every node/.style={inner sep=1pt, font=\small}
]
\node {$a$}
  child { node {$a$}
    child { node {$\square$} }
    child { node {$b$} child { node {$\square$} } }
  }
  child { node {$b$} child { node {$\square$} } };
\end{tikzpicture}
\caption{The encoding $\varphi(aabb) = a(a(\square,\, b(\square)),\,
b(\square))$ of the Dyck-1 word $aabb$. Each push letter becomes a
binary node carrying its matched pop as the right child; each pop is a
unary node; leaves $\square$ stand for the empty word.}
\label{fig:phi-example}
\end{figure}

To each well-formed word, the BParse construction of~\cite{barbot}
assigns a tree over the ranked alphabet $\Gamma$ with
$\Gamma_0 = \{\square\}$, $\Gamma_1 = \Sigma_\mathit{pop} \cup
\Sigma_\mathit{int}$, and $\Gamma_2 = \Sigma_\mathit{push}$. The
encoding $\varphi : W_\Sigma \to \mathsf{Trees}(\Gamma)$ is defined inductively by
$\varphi(\varepsilon) = \square$, $\varphi(c\,w) = c(\varphi(w))$ for
$c \in \Sigma_\mathit{int}$, and $\varphi(a\,w_1\,b\,w_2) \;=\;
a\!\left(\varphi(w_1),\; b(\varphi(w_2))\right)$ for
$a \in \Sigma_\mathit{push}$ and $b \in \Sigma_\mathit{pop}$. The map is a bijection between $W_\Sigma$ and its image
$\mathcal{T}_\Gamma = \varphi(W_\Sigma) \subseteq \mathsf{Trees}(\Gamma)$, the
\emph{BParse-shaped trees}. The reverse direction extends to a map
$\mathrm{seq} : \mathsf{Trees}(\Gamma) \to \Sigma^*$ defined as the pre-order
traversal of the tree's internal-node labels, with $\square$ leaves
treated as $\varepsilon$. On $\mathcal{T}_\Gamma$ the map agrees with
$\varphi^{-1}$; on the rest of $\mathsf{Trees}(\Gamma)$ it is many-to-one.
\Cref{fig:phi-example} shows the encoding of the Dyck-1 word $aabb$
under $\Sigma_\mathit{push} = \{a\}$, $\Sigma_\mathit{pop} = \{b\}$,
$\Sigma_\mathit{int} = \emptyset$.

\subsection{\emph{Gated}-VPL\textsuperscript{*} with A\textsuperscript{*}-based EQ}
\label{sec:vplstar}

\begin{figure}[t]
\centering
\begin{tikzpicture}[
  node distance=8mm and 14mm,
  font=\small,
  every node/.style={inner sep=2pt},
  io/.style={shape=ellipse, draw, aspect=2.2, minimum width=20mm, align=center},
  decision/.style={diamond, draw, minimum width=22mm, minimum height=10mm, align=center},
  terminal/.style={ellipse, draw, minimum width=28mm, minimum height=10mm, align=center},
  arr/.style={-{Latex[length=2mm]}, thick},
]
\node[io] (tree) {Tree $t$};
\node[decision, right=of tree] (q) {$t \in \mathcal{T}_\Gamma$?};
\node[terminal, above right=4mm and 16mm of q] (ask) {$T(\mathrm{seq}(t))$};
\node[terminal, below right=4mm and 16mm of q] (false) {$\mathit{false}$};
\draw[arr] (tree) -- (q);
\draw[arr] (q.east) -- node[sloped, above, font=\scriptsize] {yes} (ask.west);
\draw[arr] (q.east) -- node[sloped, below, font=\scriptsize] {no} (false.west);
\end{tikzpicture}
\caption{The gated membership policy of VPL\textsuperscript{*}. The
oracle returns $T(\mathrm{seq}(t))$ for BParse-shaped trees and
$\mathit{false}$ for everything else.}
\label{fig:mq-vpl-star}
\end{figure}
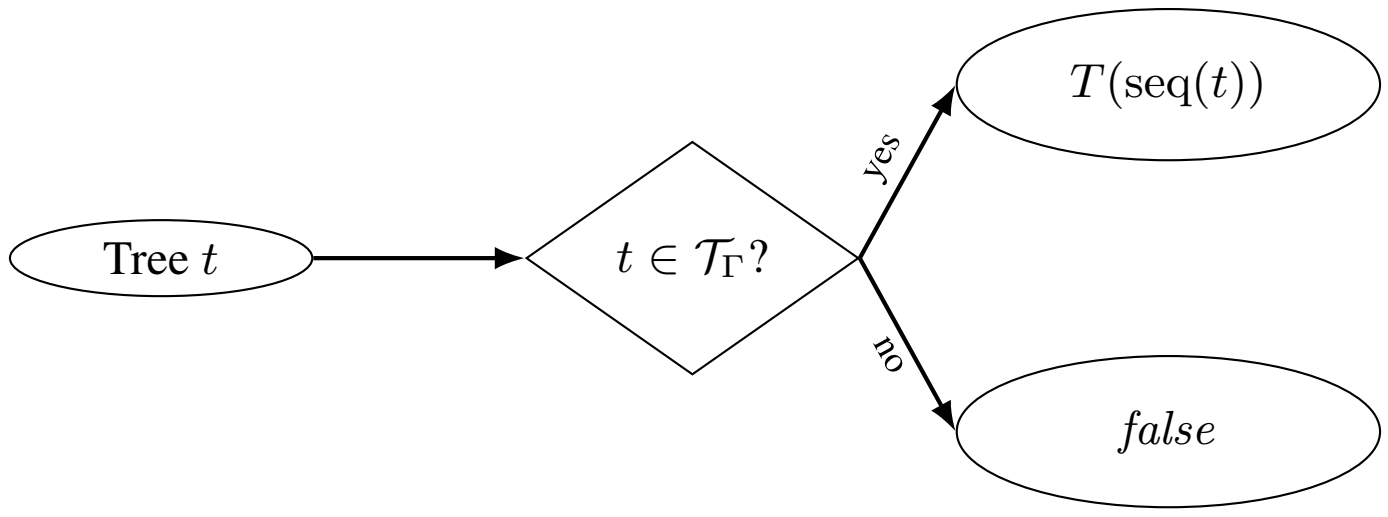

Given a binary acceptor $T : \Sigma^* \to \{0,1\}$ that realizes some
VPL $L \subseteq W_\Sigma$, the VPL\textsuperscript{*} framework of
\cite{barbot} learns a VPG for $L$ by lifting $T$ to a tree-language
target through $\varphi$ and running the tree-language extension of
L\textsuperscript{*} due to~\cite{sakakibara1992} and~\cite{drew-hogberg}
(TL\textsuperscript{*}) on top. TL\textsuperscript{*} maintains an
observation table over trees and contexts, issues membership queries on
individual trees, conjectures a deterministic tree automaton (DTA), and
refines the conjecture on counterexamples returned by an equivalence
query until convergence. Membership queries against $T$ are answered by
\[
  \mathrm{MQ}^{\mathrm{gated}}(t) \;=\;
  \begin{cases}
    T(\mathrm{seq}(t)) & \text{if } t \in \mathcal{T}_\Gamma, \\
    0 & \text{otherwise,}
  \end{cases}
\]
that is, trees outside the BParse image are gated to \emph{false}
before the target is consulted; the DTA produced by
TL\textsuperscript{*} is then post-processed into a VPG by the
construction of~\cite{barbot}. 
We refer to this as \emph{gated}-VPL\textsuperscript{*}.
\Cref{fig:mq-vpl-star} shows the
decision the gated membership oracle makes on each incoming tree. 
The equivalence query in their framework relies on
A\textsuperscript{*} search over the internal state of the RNN that
realizes $T$, which ties the algorithm to recurrent neural networks in a white-box setting.

%% file: src/shortages.tex
\section{Analysis of shortcomings}
\label{sec:short}

\subsection{\emph{Gated} policy}
\label{sec:gated-policy}

The gated policy partitions $\mathsf{Trees}(\Gamma)$ into one region that the
target sees and two that it does not. Made explicit:
\begin{itemize}
  \item \textbf{Class A}: $t \in \mathcal{T}_\Gamma$. The tree is
    BParse-shaped; the target is consulted on $\mathrm{seq}(t)$.
  \item \textbf{Class B}: $t \notin \mathcal{T}_\Gamma$ but
    $\mathrm{seq}(t) \in W_\Sigma$. The sequence is well-matched even
    though the tree is not BParse-shaped, and the BParse-shaped tree
    $\varphi(\mathrm{seq}(t))$ is a class-A tree that yields the same
    sequence.
  \item \textbf{Class C}: $t \notin \mathcal{T}_\Gamma$ and
    $\mathrm{seq}(t) \notin W_\Sigma$. The sequence is ill-matched;
    no BParse-shaped tree yields it.
\end{itemize}
Classes B and C are both gated to $\mathit{false}$ by
$\mathrm{MQ}^{\mathrm{gated}}$. For class B the gated $\mathit{false}$
is a redundancy: the class-A tree $\varphi(\mathrm{seq}(t))$ already
records $T(\mathrm{seq}(t))$, so the class-B label adds no information
the observation table does not already have. Class C is where the gate
hides information: the target may accept $\mathrm{seq}(t)$, and the
gate suppresses that signal before the learner can react to it. The
next sections lift the algorithm into a fully black-box regime
and then recover the class-C signal in two complementary ways.

\subsection{White-box equivalence}
\label{sec:pac}

The framework of \cite{barbot} relies on random sampling of sequences and A\textsuperscript{*} 
search over the internal state of a recurrent network for equivalence queries, which
constrains its application to white-box recurrent neural architectures. 

In order to enable grammar inference on arbitrary acceptors, 
while providing statistical guarantees about the learned model,
our first contribution is to
replace that oracle with \emph{PAC-oracle over trees}.
%formulation of L\textsuperscript{*}~\cite{angluin1987}. 
Fix a distribution $\mathcal{D}$ over $\mathsf{Trees}(\Gamma)$ and parameters
$\epsilon, \delta \in (0,1)$. Let $\tilde T : \mathsf{Trees}(\Gamma) \to \{0,1\}$
denote the lifted target evaluated by membership queries. Given a
conjectured DTA $A$, the oracle draws an i.i.d.\ sample of size
\[
  n(i) \;=\; \frac{1}{\epsilon}\!\left(\ln\frac{1}{\delta} + i \ln 2\right)
\]
from $\mathcal{D}$ at the $i$-th equivalence query and returns the
first $t$ on which $A(t) \neq \tilde T(t)$, or \emph{equivalent} if no
disagreement is found. A standard
analysis similar to~\cite{angluin1987} bounds the probability of returning
\emph{equivalent} on a DTA that disagrees with $\tilde T$ on more than
an $\epsilon$ fraction of $\mathcal{D}$ by $\delta$ over the entire
equivalence-query history. The target is now an opaque function
$T : \Sigma^* \to \{0,1\}$ exposing only acceptance decisions; the
rest of VPL\textsuperscript{*} runs unchanged. The same machinery,
applied with different choices of $\tilde T$, drives the
membership policies developed in the following sections.
In what follows we assume uniform sampling over $\mathsf{Trees}(\Gamma)$.
The need for sampling trees rather than sequences is explained 
later in \Cref{sec:hybrid-sampling}.

%% file: src/leakage.tex
\section{Leakage analysis on the gated learner}
\label{sec:leakage}

The gated MQ of \Cref{sec:vplstar} returns $\mathit{false}$ on every
tree outside $\mathcal{T}_\Gamma$, irrespective of what the target
would say on the corresponding sequence. When the target is in fact a
VPL this is the right choice; when it is not, the gate hides
information. Our second contribution is to instrument the gated
learner with a side oracle that quantifies what the gate hides, at
the cost of one extra target query per class-C tree and no change to
the membership semantics seen by TL\textsuperscript{*}.

\subsection{What the gate hides}
\label{sec:leakage-classes}

Recall the partition of $\mathsf{Trees}(\Gamma)$ into classes A, B, C from
\Cref{sec:gated-policy}. On class A the gated MQ already issues
$T(\mathrm{seq}(t))$ and any information the target carries on
$\mathrm{seq}(t)$ enters the observation table. On class B the
sequence $\mathrm{seq}(t)$ is in $W_\Sigma$ but the tree $t$ is
a class-A tree, so the gate's $\mathit{false}$ on class B suppresses
a duplicate of an existing class-A row rather than hiding
information. Class C is where the gate can lie. The sequence is
outside $W_\Sigma$, no class-A tree yields it, and the gate's
$\mathit{false}$ can disagree with $T(\mathrm{seq}(t))$ when the
target accepts non-well-formed sequences.
Concretely, a target that is not a VPL accepts some
$w \in \Sigma^* \setminus W_\Sigma$ and the gate masks that fact.
We call a class-C tree on which the target accepts the sequence a
\emph{leakage witness}: a sequence that the target accepts
but no VPL with the same alphabet could.

\subsection{The leakage detector}
\label{sec:leakage-detector}

\begin{figure}[t]
\centering
\begin{tikzpicture}[
  node distance=5mm and 7mm,
  font=\small,
  every node/.style={inner sep=2pt},
  io/.style={shape=ellipse, draw, minimum width=12mm, align=center},
  decision/.style={diamond, draw, aspect=2.2, inner sep=0pt, align=center},
  terminal/.style={ellipse, draw, minimum width=10mm, minimum height=7mm, align=center},
  arr/.style={-{Latex[length=2mm]}, thick},
]
% Main horizontal flow
\node[io] (tree) {Tree $t$};
\node[decision, right=of tree] (q1) {$t \in \mathcal{T}_\Gamma$?};
\node[decision, right=14mm of q1] (q2) {$\mathrm{seq}(t) \in W_\Sigma$?};
\node[decision, right=14mm of q2] (q3) {target accepts?};
\node[terminal, right=of q3, draw=red!70!black, very thick, fill=red!8] (leak) {\textbf{leakage}};
% Above-line terminals
\node[terminal, above=8mm of q1] (ask) {$T(\mathrm{seq}(t))$};
\node[terminal, above=8mm of q2] (nolk1) {no leakage};
\node[terminal, above=8mm of q3] (nolk2) {no leakage};
% Red highlight around the analysis subgraph
\begin{scope}[on background layer]
  \node[draw=red!70!black, thick, rounded corners, inner sep=3mm,
        fit=(q2)(q3)(nolk1)(nolk2)(leak),
        label={[red!70!black, font=\scriptsize, anchor=south west]south west:\textit{leakage analysis}}] {};
\end{scope}
% Arrows
\draw[arr] (tree) -- (q1);
\draw[arr] (q1) -- node[right, font=\scriptsize] {yes} (ask);
\draw[arr] (q1) -- node[above, font=\scriptsize, align=center]
  {no,\\$\mathit{false}$} (q2);
\draw[arr] (q2) -- node[right, font=\scriptsize] {yes} (nolk1);
\draw[arr] (q2) -- node[above, font=\scriptsize] {no} (q3);
\draw[arr] (q3) -- node[right, font=\scriptsize] {no} (nolk2);
\draw[arr] (q3) -- node[above, font=\scriptsize] {yes} (leak);
\end{tikzpicture}
\caption{Gated VPL\textsuperscript{*} with the leakage detector. The
main row is the original gated MQ followed by the leakage analysis
(boxed in red): a class-B filter and a target query on class C that
records a witness when the target accepts. The value returned to
TL\textsuperscript{*} is unchanged.}
\label{fig:mq-leakage}
\end{figure}
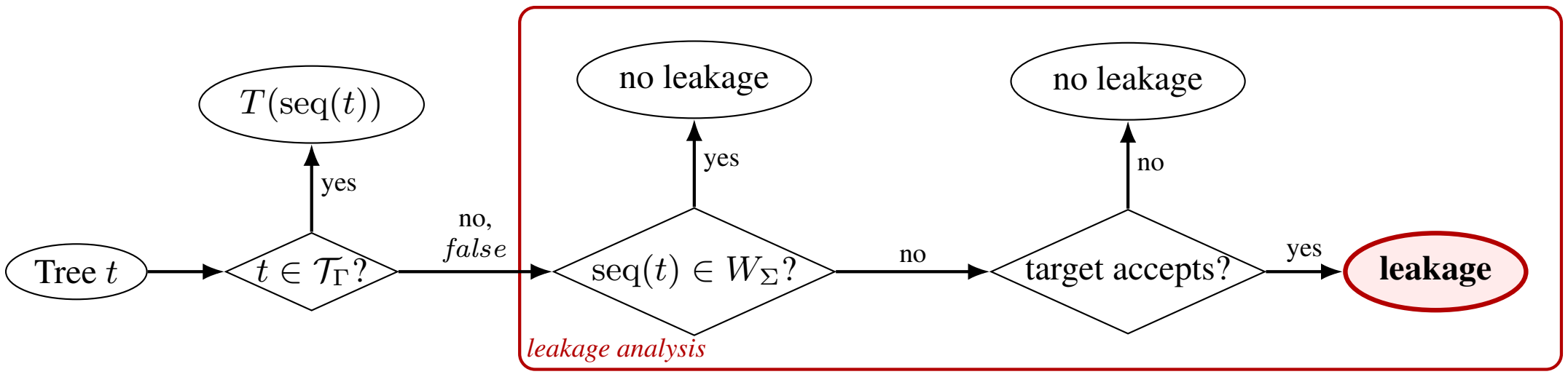

The detector is a side observer added to the gated MQ. When the
gated MQ is called on a tree $t$, the detector classifies $t$ and
acts only on class C: it issues a single extra query
$T(\mathrm{seq}(t))$ and records the pair
$(\mathrm{seq}(t), T(\mathrm{seq}(t)))$. The membership value
returned to TL\textsuperscript{*} is unchanged --- still
$\mathit{false}$ on class C --- so the observation table, the DTA,
and the extracted VPG are exactly what the gated algorithm of
\cite{barbot} would have produced. \Cref{fig:mq-leakage} shows the
augmented decision: the upper branch is the original
$\mathrm{MQ}^{\mathrm{gated}}$, and the lower branch routes the
class-A complement through an additional check that filters out
class B before issuing the target query that drives the leakage
signal.

A \emph{probe} is an instance of the lower branch's target query. A
\emph{witness} is a probe the target accepted. The probe distribution is the
distribution of class-C trees the gated learner happens to issue
during its run; it is shaped by the learner's counterexample stream,
not by an external sampler. Two consequences follow. First, a
witness is hard evidence: a sequence $w \notin W_\Sigma$ with
$T(w) = 1$. Any single such $w$ rules out the hypothesis that $T$ is
a VPL of the given alphabet, since a VPL recognizes only subsets of
$W_\Sigma$. Second, the absence of witnesses is weaker evidence:
the probes are not uniform over $\Sigma^* \setminus W_\Sigma$, so the
witness rate measured on a run does not lift to a PAC bound on the
target's acceptance rate over the non-BParse population. A
PAC-grounded estimate of that rate requires a separate stratified
sampler that draws class-C trees from a fixed distribution
independently of the learner's trajectory. The detector itself is a
witness collector, not a rate estimator.

\subsection{Cost and integration}
\label{sec:leakage-cost}

The detector adds one BParse check per non-class-A tree and one
target query per class-C tree, both linear in $|\mathrm{seq}(t)|$.
The count of gated queries is unaffected; the only extra cost
shows up as additional target calls outside the MQ counter. The
gated VPG is produced exactly as before; the detector emits a list of
named witnesses, each tagged with a malformation type --- an excess
close, an excess open, or a more general non-bracket-balanced
sequence --- and the raw counts of probes and witnesses issued during
the run. The two outputs are complementary: the VPG is the gated
learner's formal answer about the target on $W_\Sigma$, and the
witness list is a concrete refutation, when one exists, of the
assumption that $T$ is a VPL at all.

%% file: src/hybrid.tex
\section{The hybrid membership policy}
\label{sec:hybrid}

The leakage detector of \Cref{sec:leakage} records the target's
behaviour outside $\mathcal{T}_\Gamma$ but does not let the learner
use it: every class-C tree is still gated to $\mathit{false}$ before
TL\textsuperscript{*} sees the label. Our third contribution lifts
the gate where it can be lifted safely. We define a new MQ policy
that returns $T(\mathrm{seq}(t))$ on class A (as the gate already
does), $\mathit{false}$ on class B (the duplicate-suppression case),
and $T(\mathrm{seq}(t))$ on class C (where the gate previously
suppressed information). The learner's MQ stream now carries the
class-C signal directly; the witness log of \Cref{sec:leakage} is
still produced as a side effect.
%
%\subsection{Definition}
%\label{sec:hybrid-definition}
%
The \emph{hybrid} membership oracle is
\[
  \mathrm{MQ}^{\mathrm{hybrid}}(t) \;=\;
  \begin{cases}
    T(\mathrm{seq}(t)) & \text{if } t \in \mathcal{T}_\Gamma
      \text{ or } \mathrm{seq}(t) \notin W_\Sigma, \\
    \mathit{false} & \text{otherwise (class B).}
  \end{cases}
\]
The class-A and class-B branches coincide with $\mathrm{MQ}^{\mathrm{gated}}$.
The class-C branch is the new one: instead of returning
$\mathit{false}$ before consulting the target, the oracle asks the
target on the sequence and returns its answer. \Cref{fig:mq-hybrid}
illustrates the decision: the upper branch is unchanged from
\Cref{sec:vplstar}, and the lower branch now splits on whether the
sequence is in $W_\Sigma$; only class B falls through to $\mathit{false}$,
while class C feeds the target's response back to
TL\textsuperscript{*}.

\begin{figure}[t]
\centering
\begin{tikzpicture}[
  node distance=4mm and 5mm,
  font=\footnotesize,
  every node/.style={inner sep=1.5pt},
  io/.style={shape=ellipse, draw, minimum width=10mm, align=center},
  decision/.style={diamond, draw, aspect=2.4, inner sep=0pt, align=center},
  terminal/.style={ellipse, draw, minimum width=14mm, minimum height=6.5mm, align=center},
  arr/.style={-{Latex[length=1.8mm]}, thick},
]
% Main horizontal flow
\node[io] (tree) {Tree $t$};
\node[decision, right=of tree] (q1) {$t \in \mathcal{T}_\Gamma$?};
\node[decision, right=9mm of q1] (q2) {$\mathrm{seq}(t) \in W_\Sigma$?};
\node[decision, right=9mm of q2] (q3) {target accepts?};
\node[terminal, right=of q3, draw=green!50!black, very thick, fill=green!8,
      minimum width=18mm, align=center]
  (trueC) {$\mathit{true}$ + witness\\[0pt]{\scriptsize\itshape (not used for VPG)}};
% Above-line terminals
\node[terminal, above=6mm of q1] (askA) {$T(\mathrm{seq}(t))$};
\node[terminal, above=6mm of q2] (falseB) {$\mathit{false}$};
\node[terminal, above=6mm of q3] (falseC) {$\mathit{false}$};
% Arrows
\draw[arr] (tree) -- (q1);
\draw[arr] (q1) -- node[right, font=\scriptsize] {yes} (askA);
\draw[arr] (q1) -- node[above, font=\scriptsize] {no} (q2);
\draw[arr] (q2) -- node[right, font=\scriptsize] {yes (B)} (falseB);
\draw[arr] (q2) -- node[above, font=\scriptsize] {no (C)} (q3);
\draw[arr] (q3) -- node[right, font=\scriptsize] {no} (falseC);
\draw[arr] (q3) -- node[above, font=\scriptsize] {yes} (trueC);
\end{tikzpicture}
\caption{The hybrid membership policy. Class A (top branch) behaves
as in the original gated policy. Class B is forced to $\mathit{false}$
to suppress duplicates of class-A queries. Class C queries the target
directly; an accepting target produces both a $\mathit{true}$ value
fed back to TL\textsuperscript{*} and a leakage witness in the sense
of \Cref{sec:leakage}. The $\mathit{true}$ class-C rows (boxed in
green) shape the DTA but are unreachable from the extracted VPG.}
\label{fig:mq-hybrid}
\end{figure}
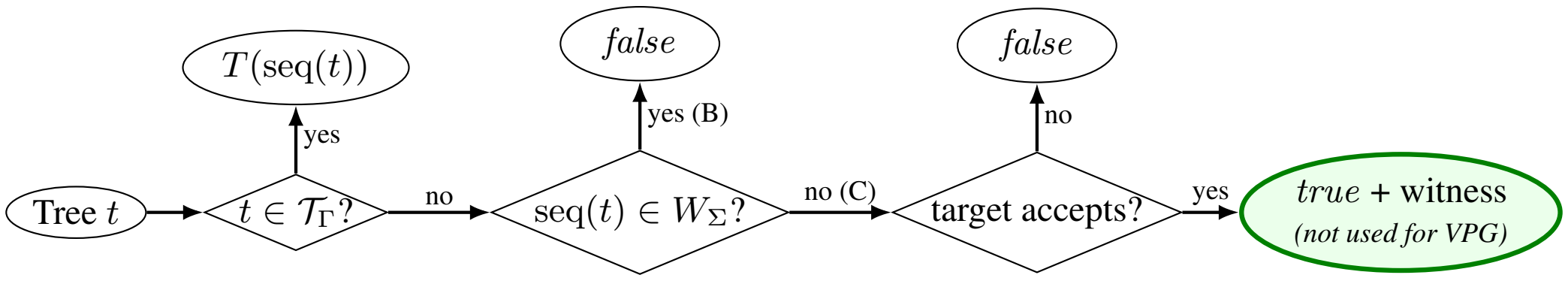

\begin{figure}[t]
\centering
\begin{tabular}{c@{\hspace{14mm}}c}
\begin{tikzpicture}[
  level distance=8mm, level 1/.style={sibling distance=18mm},
  level 2/.style={sibling distance=10mm}, level 3/.style={sibling distance=6mm},
  every node/.style={inner sep=1pt, font=\small}
]
\node {$a$}
  child { node {$a$}
    child { node {$b$} child { node {$\square$} } }
    child { node {$\square$} }
  }
  child { node {$b$} child { node {$\square$} } };
\end{tikzpicture}
&
\begin{tikzpicture}[
  level distance=8mm, level 1/.style={sibling distance=18mm},
  level 2/.style={sibling distance=10mm}, level 3/.style={sibling distance=6mm},
  every node/.style={inner sep=1pt, font=\small}
]
\node {$a$}
  child { node {$a$}
    child { node {$\square$} }
    child { node {$b$} child { node {$\square$} } }
  }
  child { node {$\square$} };
\end{tikzpicture}
\\[2mm]
{\small (a) Class B: $\mathrm{seq}(t) = aabb \in W_\Sigma$}
&
{\small (b) Class C: $\mathrm{seq}(t) = aab \notin W_\Sigma$}
\end{tabular}
\caption{Two malformed trees over the Dyck-1 alphabet, neither in
$\mathcal{T}_\Gamma$. (a) Flipping the inner $a$'s children in the
encoding of \Cref{fig:phi-example}: the right child should be the
matched $b$ pop but is $\square$, so $t \notin \mathcal{T}_\Gamma$,
yet $\mathrm{seq}(t) = aabb$ is well-matched. The class-A tree
$\varphi(aabb)$ already records $T(aabb)$, so the hybrid gate fires
on this tree. (b) The root's right child is $\square$, breaking
BParse; the pre-order $\mathrm{seq}(t) = aab$ is unbalanced, so no
class-A tree yields it, and the hybrid policy returns $T(aab)$
rather than $\mathit{false}$.}
\label{fig:raw-hyb-example}
\end{figure}
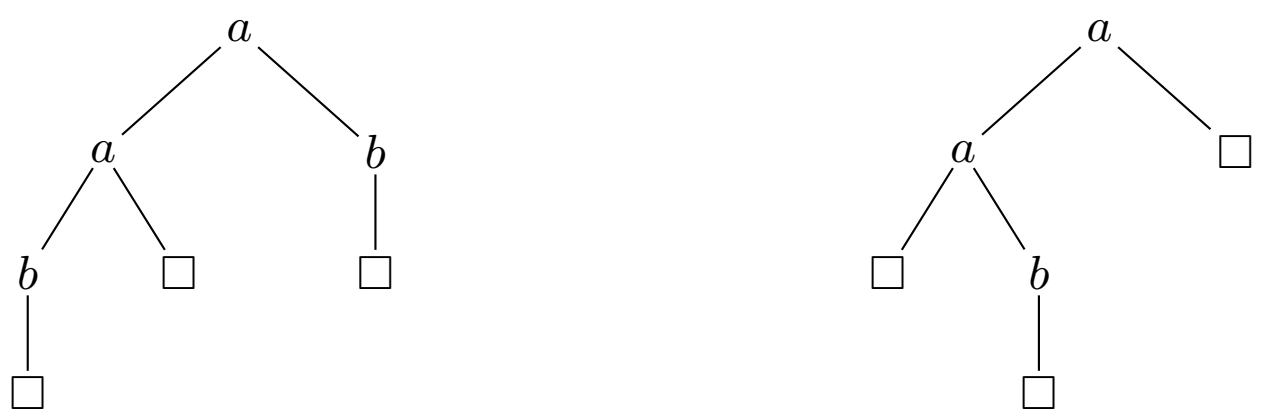

\subsection{Raw gate}

A natural alternative would be to drop the gate entirely and run
$\mathrm{MQ}(t) = T(\mathrm{seq}(t))$ on every tree. We call
this the \emph{raw} policy. The raw policy is what tempts the
reader once the leakage detector has shown that class-C information
is real; the issue is class B. Two class-B trees $t, t'$ with
$\mathrm{seq}(t) = \mathrm{seq}(t')$ carry identical labels
under raw, so the observation table records the same target
information twice. Worse, the same label is also carried by the
class-A tree $\varphi(\mathrm{seq}(t)) \in \mathcal{T}_\Gamma$ that
yields the same sequence. The raw policy therefore floods the
table with duplicates and changes the equivalence semantics on
$\mathcal{T}_\Gamma$: states that the gated learner would have
identified become distinguishable by class-B contexts whose only
content is a class-A row in disguise. The hybrid policy keeps the
gate on class B precisely to avoid this trap; the duplicate-suppression
property of the gate is preserved exactly where it is needed.
\Cref{fig:raw-hyb-example} shows examples of class B and C trees. 

\subsection{Why sample trees and not sequences}
\label{sec:hybrid-sampling}

Under the gated policy, the PAC oracle could in principle draw words
$w \in W_\Sigma$ from a sequence distribution and lift through
$\varphi$; this works because $\varphi$ is a bijection between
$W_\Sigma$ and $\mathcal{T}_\Gamma$, and the gate forces a label of
$\mathit{false}$ on every tree outside $\mathcal{T}_\Gamma$ before
the target is consulted. The hybrid policy breaks that argument.
On class C the MQ reads the target on $\mathrm{seq}(t)$, and
$\mathrm{seq}$ is many-to-one on $\mathsf{Trees}(\Gamma) \setminus
\mathcal{T}_\Gamma$: distinct trees with the same yield receive the
same target label but are distinct rows in the observation table.
There is no canonical inverse from sequences to trees outside the
bijection's domain $W_\Sigma \leftrightarrow \mathcal{T}_\Gamma$, so
the equivalence oracle has to sample trees directly. We adopt tree sampling throughout the paper, both to make
the gated and hybrid algorithms run on a single sampling regime and
to keep the comparison in \Cref{sec:experiments} on equal footing.

\subsection{On-$\mathcal{T}_\Gamma$ equivalence with the gated learner}
\label{sec:hybrid-equivalence}

The hybrid MQ differs from the gated MQ only on class C. Two
questions follow. Does the hybrid learner terminates? When it does,
does its VPG agree with the gated VPG on $\mathcal{T}_\Gamma$?

Termination is ensured by TL\textsuperscript{*} \cite{drew-hogberg} 
if the language $L$ of the target sequence acceptor $T$ is equal to
$\{\mathrm{seq}(t) \mid M(t) = 1,\ t\in\mathsf{Trees}(\Gamma)\}$, 
for some tree automaton acceptor $M$ over $\Gamma$.
Therefore, the class of sequence acceptors learned by the hybrid policy is
larger than that of the gated policy.

Let us focus now on agreement on $\mathcal{T}_\Gamma$. %is the more interesting question.
The VPG produced by the hybrid policy, is obtained from the learned
DTA via the same \texttt{nta2vpg} construction used in
VPL\textsuperscript{*}: read the BParse-pattern transitions
$a(\cdot,\, b(\cdot))$ off the DTA and emit one VPG rule per pattern.
The construction depends only on the DTA's behavior on
$\mathcal{T}_\Gamma$. Empirically~\Cref{sec:experiments}: on every one of the
$18$ targets in the paper the VPGs result of the hybrid and gated learners agree on every
BParse-shaped tree in a $2{,}000$-tree sample. The DTA obtained by the former can
nonetheless be much larger than the DTA of the gated one on a non-VPL target
--- up to $53$ states and $259$ NTA transitions on the bidirectional
Dyck-1 transformer \texttt{mol\_bidir}, and $21$ states and $429$
transitions on the synthetic all-accepting target, versus $2$ and
$2$--$3$ for gated --- but the extra states are unreachable from the
VPG built by \texttt{nta2vpg}, because the wrapper constructs
trees from sequences via $\varphi$ and every $\varphi(w)$ lies in
$\mathcal{T}_\Gamma$. The contribution of the
hybrid policy is therefore to fold the leakage signal of
\Cref{sec:leakage} into the learner's normal trajectory, amplifying
the witness collection without any new hyperparameter, while
preserving the VPG on $\mathcal{T}_\Gamma$.

%% file: src/experiments.tex
\section{Experiments}
\label{sec:experiments}

We evaluate the leakage detector of \Cref{sec:leakage} and the
hybrid policy of \Cref{sec:hybrid} on three target families. E1
(\Cref{sec:experiments-e1}) checks that on true VPL targets the
hybrid pipeline produces no witnesses and reduces to the gated one.
E2 (\Cref{sec:experiments-e2}) uses two synthetic non-VPL targets
to exhibit the witness signal and the on-$\mathcal{T}_\Gamma$
equivalence between the obtained VPGs. E3
(\Cref{sec:experiments-e3}) applies the same pipeline to a
bidirectional transformer trained on $\mathrm{Dyck}_1$ from~\cite{molinolo}.

\subsection{Setup}
\label{sec:experiments-setup}

Let's recall that a probe is a membership query on a class-C tree --- a tree
outside $\mathcal{T}_\Gamma$ whose sequence is also outside $W_\Sigma$.
Every probe issues one target query $T(\mathrm{seq}(t))$, whose
result both enters the observation table (under the hybrid policy)
and is logged. A probe is a witness if the target accepts on
its sequence form. Class-B queries are silent: they are forced to
$\mathit{false}$ by the gate and are not logged, since
$T(\mathrm{seq}(t))$ is already recorded by the class-A tree
$\varphi(\mathrm{seq}(t))$ that yields the same sequence. 
%Probes collected during a run come from
%the learner's counterexample trajectory rather than a fixed
%distribution, so we report probe and witness counts as raw run
%statistics.

We compare two pipelines on every target: gated VPL\textsuperscript{*}
%(the original Barbot algorithm) 
and hybrid VPL\textsuperscript{*}.
For each pipeline we record the DTA, the corresponding VPG (built by
\texttt{nta2vpg}), the membership-query count, and the class-C
probe and witness counts. To compare the two extracted grammars as
languages we draw a fixed sample of $2{,}000$ random trees (max depth
$8$) from $\mathsf{Trees}(\Gamma)$ and check, for each tree $t$ with
$t \in \mathcal{T}_\Gamma$, whether the gated ($g$) and hybrid ($h$) VPGs return
the same acceptance bit on $\mathrm{seq}(t)$. The fraction of
agreement is reported as $g \equiv h$ on $\mathcal{T}_\Gamma$ in each
table below; $100\%$ means the two grammars accept exactly the same
BParse-shaped strings on the sample. All E1 and E2 runs use seed
$71$ and $\epsilon = \delta = 0.005$. E3 uses seed $42$ and 
parameters $\epsilon = \delta = 0.05$, $\mathit{max\_depth} = 5$,
$n_\mathit{sample} = 200$ due to the per-query cost of the
transformer forward pass.

\subsection{E1: VPG targets}
\label{sec:experiments-e1}

The 15 visibly pushdown targets of \cite{barbot} are each a VPL by
construction, so on every class-C tree the target rejects the sequence.
The hybrid MQ therefore returns $\mathit{false}$ on every class-C
tree, exactly as the gated MQ would, and the two trajectories
coincide. \Cref{tab:e1} reports DTA, rule, MQ, probe, and witness
counts. The values are identical between the gated and hybrid
pipelines on every target; the table reports a single row per target.
Across $24{,}228$ class-C probes the target rejected every one. The
hybrid policy collapses to the gated one as predicted; the entries
of \Cref{tab:e1} are therefore also the gated baseline for E1.

\begin{table}[t]
\centering
\footnotesize
\begin{tabular}{l c c c c c c}
\toprule
Target & States & Rules & MQs & Probes & Witnesses & $g \equiv h$ on $\mathcal{T}_\Gamma$ \\
\midrule
L1  & 2 & 2 & $1{,}187$ & $741$   & $0$ & 100\% \\
L2  & 2 & 5 & $2{,}227$ & $1{,}693$ & $0$ & 100\% \\
L3  & 2 & 3 & $2{,}448$ & $2{,}165$ & $0$ & 100\% \\
L4  & 2 & 5 & $1{,}500$ & $1{,}155$ & $0$ & 100\% \\
L5  & 2 & 5 & $1{,}975$ & $1{,}664$ & $0$ & 100\% \\
L6  & 2 & 7 & $2{,}116$ & $1{,}594$ & $0$ & 100\% \\
L7  & 3 & 3 & $2{,}182$ & $1{,}676$ & $0$ & 100\% \\
L8  & 1 & 1 & $1{,}064$ & $898$   & $0$ & 100\% \\
L9  & 2 & 1 & $2{,}464$ & $2{,}172$ & $0$ & 100\% \\
L10 & 3 & 7 & $4{,}873$ & $4{,}387$ & $0$ & 100\% \\
L11 & 4 & 3 & $3{,}547$ & $2{,}668$ & $0$ & 100\% \\
L12 & 1 & 1 & $1{,}062$ & $812$   & $0$ & 100\% \\
L13 & 2 & 5 & $1{,}319$ & $789$   & $0$ & 100\% \\
L14 & 2 & 4 & $1{,}465$ & $1{,}101$ & $0$ & 100\% \\
L15 & 2 & 6 & $1{,}226$ & $713$   & $0$ & 100\% \\
\midrule
Total & --- & --- & $30{,}655$ & $24{,}228$ & $\mathbf{0}$ & --- \\
\bottomrule
\end{tabular}
\caption{Gated and hybrid VPL\textsuperscript{*} on the 15 visibly
pushdown targets of \cite{barbot}. State, rule, and MQ counts are
identical between the two pipelines, and the extracted VPGs accept
the same trees in $\mathcal{T}_\Gamma$ on the $2{,}000$-tree sample.
No probe produced a witness.}
\label{tab:e1}
\end{table}

\subsection{E2: non-VPG targets}
\label{sec:experiments-e2}

Two synthetic targets sit outside the VPL class in qualitatively
different ways. Both use the alphabet
$\Sigma_\mathit{push} = \{(\}$, $\Sigma_\mathit{pop} = \{)\}$; the
first adds the internal symbol $a$. The first target (\emph{E2,
all-accepting}) returns $\mathit{true}$ on every $w \in \Sigma^*$,
including every non-well-matched word. The second (\emph{E2b}) accepts
every well-matched Dyck-1 word and every sequence of the form $) w$
for $w$ well-matched; its restriction to $W_\Sigma$ is exactly
$\mathrm{Dyck}_1$. \Cref{tab:e2} reports both pipelines on both
targets.

\begin{table}[t]
\centering
\footnotesize
\begin{tabular}{l c c c c c c c}
\toprule
Target & Pol. & States & Rules & MQs & Probes & Witnesses & $g \equiv h$ on $\mathcal{T}_\Gamma$ \\
\midrule
E2 (all-accepting)  & $g$ & $2$  & $3$   & $1{,}117$ & $684$   & $684$   & --- \\
E2 (all-accepting)  & $h$ & $21$ & $429$ & $45{,}051$ & $33{,}478$ & $33{,}478$ & $100\%$ \\
\addlinespace
E2b (Dyck-1 with $)$ prefix) & $g$ & $2$  & $2$  & $1{,}209$ & $708$   & $134$   & --- \\
E2b (Dyck-1 with $)$ prefix) & $h$ & $12$ & $20$ & $13{,}491$ & $8{,}576$ & $1{,}777$ & $100\%$ \\
\bottomrule
\end{tabular}
\caption{Gated ($g$) versus hybrid ($h$) policies on the
two synthetic non-VPG targets. The DTA of the hybrid policy grows substantially on
both targets, but on the $2{,}000$-tree sample the
VPGs accept exactly the same BParse-shaped strings.}
\label{tab:e2}
\end{table}

On both targets the hybrid pipeline finds many more witnesses than
the gated detector. On the all-accepting target every class-C probe
is a witness, and hybrid issues $33{,}478$ such probes versus
$684$ under gated; on E2b the witnesses are concentrated in the
\textit{excess close} malformation class --- $1{,}777$ versus $134$,
with zero \textit{excess open} witnesses in either policy, which is
the structural fingerprint of the deviation. The hybrid DTAs are
also much larger ($21$ states and $429$ NTA transitions on E2,
$12$ and $20$ on E2b) than the gated ones ($2$ and $3$, and $2$ and $2$,
respectively). The gap matters at the DTA level but not at the VPG
level: the rightmost column shows that the two extracted grammars
label every BParse-shaped tree in the sample identically, so the
gated and hybrid VPGs are equivalent on the test dataset of well-formed strings. The extra hybrid
states and rules are unreachable from the VPG built by
\texttt{nta2vpg}, which constructs trees from sequences via
$\varphi$ and therefore only consults the class-A portion of the
DTA.

\subsection{E3: neural classifiers}
\label{sec:experiments-e3}

We finally apply the pipeline to a bidirectional transformer
classifier (\texttt{mol\_bidir}) trained on $\mathrm{Dyck}_1$ by
\cite{molinolo}.\footnote{We also evaluated the causal counterpart
from the same author but it rejected every non-empty input in our
sample; both pipelines correctly extracted a 1-state DTA and the
target offered no contrast between gated and hybrid, so we drop it
from the headline numbers.} The wrapper exposes the model as a
binary acceptor of strings over $\{(, )\}$. \Cref{tab:e3} reports the run under both policies.

\begin{table}[t]
\centering
\footnotesize
\begin{tabular}{l c c c c c c c}
\toprule
Model & Pol. & States & Rules & MQs & Probes & Witnesses & $g\equiv h$ on $\mathcal{T}_\Gamma$ \\
\midrule
\texttt{mol\_bidir} & $g$ & $2$  & $2$   & $136$      & $78$      & $33$    & --- \\
\texttt{mol\_bidir} & $h$ & $53$ & $259$ & $13{,}594$ & $12{,}001$ & $9{,}535$ & $100\%$ \\
\bottomrule
\end{tabular}
\caption{Gated ($g$) versus hybrid ($h$) policies on
\cite{molinolo}'s bidirectional $\mathrm{Dyck}_1$ transformer.
%(seed $42$, smoke parameters $\epsilon = \delta = 0.05$, $\mathit{max\_depth} = 5$). 
The hybrid pipeline produces a
substantially larger DTA and many more witnesses, yet the extracted
VPGs accept exactly the same trees in $\mathcal{T}_\Gamma$ on the
sample.}
\label{tab:e3}
\end{table}

The pipelines diverge sharply on what they see
outside $\mathcal{T}_\Gamma$: the gated learner stops after $136$
MQs and $78$ class-C probes, of which $33$ are accepted by the
model; the hybrid learner uses those accepts as new table rows, its
trajectory extends, and it ends up issuing $12{,}001$ class-C probes
($154\times$ more than gated) with $9{,}535$ witnesses ($289\times$
more, balanced between $4{,}464$ excess-close and $5{,}071$
excess-open accepted sequences). Every one of those extra accepts
is invisible to the gated pipeline by construction --- the gate
suppresses the label and the gated trajectory never visits the
corresponding region of $\mathsf{Trees}(\Gamma)$. 

On-$\mathcal{T}_\Gamma$ agreement evaluated empirically on the $n_{samples}$ samples, 
showed that the VPL learned by both gated and hybrid policies, 
is the one of the $\mathrm{Dyck}_1$ grammar, in which it was trained.
While the DTA obtained by the gated policy has $2$ states, 
and \texttt{nta2vpg} resulted exactly in the $\mathrm{Dyck}_1$ grammar with $2$ rules, 
the DTA produced by the hybrid policy has $53$ states and the \texttt{nta2vpg} 
construction emits $259$ rules. However, most rules are unreachable or duplicated.

%% file: src/concl.tex
\section{Conclusions}
\label{sec:concl}

We extended the VPL\textsuperscript{*} framework of~\cite{barbot}
along three axes. First, we replaced the white-box
A\textsuperscript{*} equivalence oracle with a PAC oracle that
samples trees from $\mathsf{Trees}(\Gamma)$ and queries the target
only through its binary output, so the pipeline applies to any
acceptor exposing only binary decisions; in particular it applies to
transformer acceptors, for which the original A\textsuperscript{*}
oracle is unavailable. Second, we added a leakage detector to the
gated learner: it issues one extra target query per class-C tree and
records witnesses --- sequences outside $W_\Sigma$ that the target
accepts, each of which refutes the hypothesis that the target is a
VPL of the given alphabet --- without changing the membership
semantics seen by TL\textsuperscript{*}. Third, we defined a hybrid
membership policy that lifts the gate on class C and keeps it on
class B, folding the leakage signal into the observation table.

The experiments leave three highlights:
\begin{itemize}
  \item \textbf{Hybrid extends the space of learnable targets.} On every
    target we tested, hybrid converged to a VPG; the gated pipeline
    is exactly hybrid restricted to the class-A and class-B branches,
    and inherits the same outcome on VPL targets while remaining
    silent about class C on non-VPL ones.
  \item \textbf{Gated and hybrid VPGs agree empirically.} On every
    one of the $18$ targets the two VPGs accepted the same
    BParse-shaped trees on the $2{,}000$-tree comparison sample. We
    do not prove equivalence on $\mathcal{T}_\Gamma$; the agreement
    reported is empirical and the question of whether it holds in
    general is left open.
  \item \textbf{The extracted VPG is not minimal.} On non-VPL targets,
    the experiments showcased that the hybrid policy can result in DTAs
    larger than the gated one, and the \texttt{nta2vpg} projection 
    inherits this increase in size, although most rules were useless.
    Minimizing the output is not straightforward: \cite{alur2005congruences} show that, unlike regular languages, 
    visibly pushdown languages in general admit no unique minimal automaton. 
    Reducing the extracted grammar is therefore left for future work.
\end{itemize}

%% file: references.bib
@article{angluin1987,
  author    = {Angluin, Dana},
  title     = {Learning regular sets from queries and counterexamples},
  journal   = {Information and Computation},
  volume    = {75},
  number    = {2},
  pages     = {87--106},
  year      = {1987}
}

@article{drew-hogberg,
author = {Drewes, Frank and H\"{o}gberg, Johanna},
title = {Query Learning of Regular Tree Languages: How to Avoid Dead States},
year = {2007},
issue_date = {February 2007},
publisher = {Springer-Verlag},
address = {Berlin, Heidelberg},
volume = {40},
number = {2},
issn = {1432-4350},
doi = {10.1007/s00224-005-1233-3},
journal = {Theor. Comp. Sys.},
month = feb,
pages = {163–185},
numpages = {23}
}

@misc{jia2024vstar,
  author       = {Jia, Xiaodong and Tan, Gang},
  title        = {{V-Star}: Learning Visibly Pushdown Grammars from Program Inputs},
  year         = {2024},
  eprint       = {2404.04201},
  archivePrefix = {arXiv},
  primaryClass = {cs.PL},
  url          = {https://arxiv.org/abs/2404.04201}
}

@inproceedings{alur2005congruences,
  author    = {Alur, Rajeev and Kumar, Viraj and Madhusudan, P. and Viswanathan, Mahesh},
  title     = {Congruences for Visibly Pushdown Languages},
  booktitle = {Automata, Languages and Programming, 32nd International Colloquium (ICALP)},
  series    = {Lecture Notes in Computer Science},
  volume    = {3580},
  pages     = {1102--1114},
  year      = {2005}
}

@inproceedings{barbot,
  author    = {Barbot, Beno{\^i}t and Bollig, Benedikt and Finkel, Alain and Haddad, Serge
               and Khmelnitsky, Igor and Leucker, Martin and Neider, Daniel and Roy, Rajarshi
               and Ye, Lina},
  title     = {Extracting Context-Free Grammars from Recurrent Neural Networks
               using Tree-Automata Learning and {A*} Search},
  booktitle = {Proceedings of the 15th International Conference on Grammatical Inference},
  series    = {Proceedings of Machine Learning Research},
  volume    = {153},
  pages     = {113--129},
  year      = {2021}
}

@inproceedings{alur2004,
  title={Visibly pushdown languages},
  author={Rajeev Alur and P. Madhusudan},
  booktitle={Symposium on the Theory of Computing},
  year={2004},
}

@inproceedings{mayr2018,
  author    = {Mayr, Franz and Yovine, Sergio},
  title     = {Regular Inference on Artificial Neural Networks},
  booktitle = {International Cross-Domain Conference for Machine Learning and Knowledge Extraction
               (CD-MAKE)},
  series    = {Lecture Notes in Computer Science},
  pages     = {350--369},
  year      = {2018}
}

@inproceedings{weiss2018,
  title = 	 {Extracting Automata from Recurrent Neural Networks Using Queries and Counterexamples},
  author =       {Weiss, Gail and Goldberg, Yoav and Yahav, Eran},
  booktitle = 	 {Proceedings of the 35th International Conference on Machine Learning},
  pages = 	 {5247--5256},
  year = 	 {2018},
  editor = 	 {Dy, Jennifer and Krause, Andreas},
  volume = 	 {80},
  series = 	 {Proceedings of Machine Learning Research},
}

@article{sakakibara1992,
  author    = {Sakakibara, Yasubumi},
  title     = {Efficient learning of context-free grammars from positive structural examples},
  journal   = {Information and Computation},
  volume    = {97},
  number    = {1},
  pages     = {23--60},
  year      = {1992}
}

@thesis{molinolo,
  author      = {Molinolo De Ferrari, Mat{\'i}as},
  title       = {Exploring Attention Patterns and Neural Activations in
                 Transformer Architectures for Sequence Classification in
                 Context-Free Grammars},
  type        = {Bachelor's thesis},
  institution = {Universidad ORT Uruguay},
  school      = {Universidad ORT Uruguay},
  year        = {2024},
  url   = {https://sisbibliotecas.ort.edu.uy/cgi-bin/koha/opac-retrieve-file.pl?id=3dcb1c0fb4357e634cd89e836c9de4b7}
}
